\documentclass[letterpaper,journal]{IEEEtran}
\usepackage[utf8]{inputenc}
\usepackage{booktabs}
\usepackage{tabularx}
\usepackage{amsmath,amsfonts}
\usepackage{algorithmic}
\usepackage{algorithm}
\usepackage{array}
\usepackage[caption=false,font=normalsize,labelfont=sf,textfont=sf]{subfig}
\usepackage{textcomp}
\usepackage{stfloats}
\usepackage{url}
\usepackage{verbatim}
\usepackage{graphicx}
\graphicspath{{figures/}}
\usepackage{cite}
\usepackage{multirow}
\usepackage{makecell}
\newcolumntype{C}{>{\centering\arraybackslash}X}

\newcolumntype{M}[1]{>{\centering\arraybackslash}m{#1}}

\renewcommand{\arraystretch}{1.8}

\begin{document}

\title{FasTac: A Curved Multispectral Vision-Based Tactile Sensor for High-Speed High-Precision 3D Shape and Force Perception}

\author{Xiaofan Lu, Kaiji Huang, Jiahui Chen, Yuankai Lin, Hua Yang, and Zhouping Yin%
\thanks{This work was supported by the Joint Funds of National Natural Science Foundation of China under Grant U22A20208. (Corresponding author: Hua Yang.)}%
\thanks{The authors are with the State Key Laboratory of Intelligent Manufacturing Equipment and Technology, School of Mechanical Science and Engineering, Huazhong University of Science and Technology, Wuhan 430074, China. Hua Yang is also with the National Innovation Institute of Digital Design and Manufacturing, Wuhan 430206, China (e-mail: huayang@hust.edu.cn).}%
}

\IEEEaftertitletext{%
\begin{center}
\footnotesize
This work has been submitted to the IEEE for possible publication. Copyright may be transferred without notice, after which this version may no longer be accessible.
\end{center}
}

\maketitle

\begin{abstract}
Curved tactile fingertips for dexterous manipulation must resolve fine contact geometry, distinguish normal and tangential loads, and capture transient signals. Existing curved vision-based tactile sensors struggle to combine accurate 3D reconstruction, three-axis force estimation, and high-speed processing in a compact form. This article presents FasTac, a curved vision-based tactile sensor integrating multispectral photometric stereo, dynamic-convolution force estimation, and hardware acceleration on a field-programmable gate array (FPGA). Single-image-sensor simultaneous multispectral imaging provides spatially aligned observations for robust surface normal estimation, followed by boundary-prior fast Poisson depth reconstruction. HyperForce uses position-aware dynamic convolution to model the spatially nonuniform mechanical response of curved elastomers and estimate three-axis forces. The complete image-to-$F_z$ pipeline is deployed on an FPGA. Experiments show that near-infrared (NIR) illumination and the boundary prior decrease depth mean absolute error (MAE) from 0.2730 mm to 0.0415 mm; HyperForce achieves normalized mean absolute error (NMAE) values of 2.74\% and 2.39\% for normal and shear forces, respectively; and FPGA deployment shortens processing latency from 3.26 ms on the GPU to 1.09 ms. Multi-object reconstruction, feedback grasping, and vibration measurement validate fine geometric perception, stable force feedback, and dynamic contact sensing.
\end{abstract}


\section{Introduction}
\IEEEPARstart{I}{n} dexterous robotic manipulation, curved fingertips are often involved in fine manipulation, stable grasping, and dynamic contact interaction, where contacts occur over small, non-planar regions, and the contact state changes rapidly~\cite{kappassov2015tactile}. In fine manipulation~\cite{lambeta2020digit,romero2020soft}, local contact imprints contain subtle geometric features, such as small protrusions, pits, and edges, which robots must perceive with high spatial resolution. In stable grasping, external disturbances change the balance between normal loading and tangential loading, so grasp stability depends on whether the available friction can resist the tangential load, rather than on the normal force alone~\cite{kim2023barotac,yang2023tactile}. In dynamic interactions, such as vibrations or collisions, transient tactile cues may not be captured in time if the sensing and computation pipeline is too slow~\cite{andrussow2023minsight}. These contact characteristics motivate compact curved tactile fingertips that can preserve fine 3D contact geometry, distinguish normal and tangential loading, and process dynamic tactile signals with low latency.

\begin{figure}[t]
\centering
\includegraphics[width=1.0\linewidth]{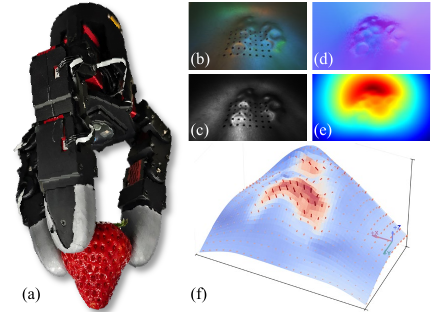}
\caption{\textbf{High-precision 3D perception of delicate objects using the FasTac sensor.}
(a) A three-fingered dexterous hand equipped with FasTac sensors grasping a strawberry. (b-c) RGB and NIR images captured by FasTac. (d) Reconstructed surface normal map of the strawberry contact region. (e) Corresponding depth map. (f) Three-axis contact force distribution estimation results visualized on the curved surface.}
\label{fig:overview}
\end{figure}

Existing tactile sensors can be broadly organized by their transduction mechanisms, including resistive, capacitive, and magnetic approaches, which respectively convert deformation into changes in resistance, capacitance, and magnetic-field distribution. Resistive tactile arrays~\cite{zheng2025large} and capacitive tactile arrays~\cite{yu2024stretchable} are effective for large-area pressure mapping, while magnetic sensors~\cite{rehan2022soft,hu2024large} provide stronger potential for three-axis force estimation. However, achieving high spatial resolution with taxel-based arrays requires smaller and denser sensing units, which can reduce signal levels while increasing wiring complexity, parasitic effects, and inter-taxel crosstalk; their discrete outputs also provide insufficient geometric information for fine 3D reconstruction~\cite{wang2023recent,hu2024large}. Although conventional transduction mechanisms can support three-axis force sensing, coupled mechanical and electrical or magnetic responses generally require specialized structures or calibration models to decouple normal and shear forces~\cite{rehan2022soft}. Curved fingertips further complicate both 3D reconstruction and three-axis force estimation because local curvature changes baseline deformation, contact orientation, boundary constraints, and effective sensitivity~\cite{wang2017flexible,zhong2025curvature}. Therefore, simultaneously achieving high-precision 3D reconstruction and reliable three-axis force estimation on a compact curved fingertip remains challenging.

Vision-based tactile sensing provides a promising approach to addressing this challenge by imaging elastomer deformation and converting images into geometry and force~\cite{yuan2017gelsight}. Planar and near-planar sensors have demonstrated high-resolution 3D reconstruction~\cite{lin20239dtact} and dense force estimation~\cite{ma2019dense}. Curved designs~\cite{gomes2020geltip,tippur2023gelsight360} improve contact adaptability and sensing coverage, but their non-planar geometry makes it difficult to achieve sufficient and uniform illumination across the sensing surface. In addition, force estimation on curved elastomers is a mechanically coupled inverse problem: local curvature or boundary constraints can change the deformation--force relationship across the surface~\cite{zhong2025curvature}. Moreover, perceiving rapid load variation like vibrations and collisions requires low-latency tactile processing, while streaming high-resolution tactile images to external processors requires image transmission, buffering, and host-side scheduling, increasing the end-to-end latency of tactile feedback~\cite{andrussow2023minsight}.

To address these challenges, this paper presents FasTac, a compact curved vision-based tactile sensor for high-speed high-precision 3D shape and force perception, as shown in Fig.~\ref{fig:overview}. FasTac uses RGB-NIR multispectral illumination with a single RGB-NIR image sensor to provide synchronized multispectral inputs for surface normal estimation on curved surfaces, followed by boundary-prior fast Poisson depth reconstruction. It further introduces HyperForce, a position-aware dynamic-convolutional model that maps depth-derived normal displacement and marker-derived tangential displacement to three-axis contact force. Finally, the image-to-$F_z$ estimation pipeline is implemented on a field-programmable gate array (FPGA) for low-latency hardware acceleration. Experiments on curved-surface shape reconstruction, friction-aware feedback grasping, and high-frequency normal-force sensing validate the sensor's capabilities for fine geometric perception, stable grasp feedback, and dynamic contact sensing.

In summary, our contributions are summarized as follows:

\begin{list}{\arabic{enumi})}{\usecounter{enumi}
\setlength{\leftmargin}{1.8em}
\setlength{\labelwidth}{1.2em}
\setlength{\labelsep}{0.4em}
\setlength{\itemsep}{0.3em}
\setlength{\topsep}{0.3em}
\setlength{\parsep}{0em}
\setlength{\partopsep}{0em}}
\item A compact curved vision-based tactile fingertip is developed by combining RGB-NIR multispectral illumination with a single RGB-NIR CMOS image sensor, providing synchronized multispectral imaging for robust 3D reconstruction in high-curvature regions.
\item A spatially adaptive three-axis force estimation method, HyperForce, is proposed to model the spatially non-uniform response of curved elastomers and estimate normal and tangential forces.
\item An FPGA image-to-$F_z$ estimation pipeline is constructed for edge deployment, enabling high-speed tactile processing.
\end{list}

Table~\ref{table:sensor_comparison} compares representative curved vision-based tactile sensors. Abbreviations are as follows: MAE, mean absolute error; NMAE, normalized mean absolute error; FEM, finite element method; Result., resultant force; Dist., distributed force; and $F_n$, $F_s$, and $F$, normal, shear, and total force, respectively. Regular and bionic identify geometric categories, and speed is the image-to-output computation rate. Because evaluation settings differ across studies, the reported values provide a capability-level comparison.

\begin{table*}[t]
\centering
\caption{\textsc{Comparison with Representative Curved Vision-Based Tactile Sensors}}
\label{table:sensor_comparison}
\scriptsize
\setlength{\tabcolsep}{4.5pt}
\renewcommand{\arraystretch}{1.0}
\begin{tabular*}{\textwidth}{@{\extracolsep{\fill}}cccccccccccc@{}}
\hline
\multicolumn{1}{c}{\multirow{2}{*}{\textbf{Sensor}}} &
\multicolumn{1}{c}{\multirow{2}{*}{\textbf{Geometry}}} &
\multicolumn{4}{c}{\textbf{Shape Reconstruction}} &
\multicolumn{5}{c}{\textbf{Force Estimation}} &
\multicolumn{1}{c}{\multirow{2}{*}{\textbf{Notes}}} \\
\cline{3-6}\cline{7-11}
& & \textbf{Method} & \textbf{Output} &
\makecell{\textbf{MAE}\\\textbf{(mm)} $\downarrow$} &
\makecell{\textbf{Speed}\\\textbf{(Hz)} $\uparrow$} &
\textbf{Method} & \textbf{Format} & \textbf{Output} &
\makecell{\textbf{NMAE}\\\textbf{(\%)} $\downarrow$} &
\makecell{\textbf{Speed}\\\textbf{(Hz)} $\uparrow$} & \\
\hline
\makecell{DenseTact\\2.0\cite{do2022densetact}} &
\makecell{Regular\\curved} & DenseNet & \makecell{Depth\\map} & 0.3633 & 25 &
DenseNet & Result. & \makecell{$F_x,F_y,F_z$\\$T_x,T_y,T_z$} & $F$: 2.93 & -- &
\makecell{6D force sensing;\\large depth error.} \\ \hline

Insight\cite{sun2022soft} &
\makecell{Bionic\\curved} & -- & -- & -- & -- &
ResNet & Dist. & $F_x,F_y,F_z$ & $F$: 4.00 & 10 &
\makecell{Three-axis force map;\\no depth reconstruction.} \\ \hline

\makecell{GelStereo\\BioTip\cite{cui2023gelstereo}} &
\makecell{Regular\\curved} & \makecell{Stereo\\vision} & \makecell{3D\\coordinates} & 0.2730 & 30 &
-- & -- & -- & -- & -- &
\makecell{Direct 3D acquisition;\\no force estimation.} \\ \hline

SoftBubble\cite{peng20243d} &
\makecell{Regular\\curved} & \makecell{RGB-D +\\optical flow} & \makecell{3D\\coordinates} & -- & -- &
FEM & Dist. & $F_x,F_y,F_z$ & $F$: 7.75 & 1--2 &
\makecell{Data-efficient;\\low force speed.} \\ \hline

GelSplitter3D\cite{lin20253d} &
\makecell{Bionic\\curved} & \makecell{Photometric\\stereo} & \makecell{Depth\\map} & \textbf{0.0406} & 30 &
-- & -- & -- & -- & -- &
\makecell{High depth accuracy;\\no force estimation.} \\ \hline

FasTac &
\makecell{Bionic\\curved} & \makecell{Photometric\\stereo} & \makecell{Depth\\map} & 0.0415 & \textbf{1022} &
\makecell{Dynamic\\convolution} & Dist. & $F_x,F_y,F_z$ &
\textbf{\makecell{$F_n$: 2.74\\$F_s$: 2.39}} & \textbf{918} &
\makecell{\textbf{Shape and three-axis force;}\\\textbf{high-speed perception.}} \\
\hline
\end{tabular*}
\end{table*}

\section{Related Work}

\subsection{Vision-Based Tactile 3D Reconstruction}
Existing reconstruction methods mainly include photometric stereo~\cite{woodham1979photometric,yuan2017gelsight}, stereo vision~\cite{lu2025stereotactip} and optical flow~\cite{zhang2022deltact,zhang2023gelflow}. Among them, GelSight-like photometric-stereo sensors recover surface normals from multi-directional illumination and then integrate them into depth maps, which is effective for fine texture reconstruction~\cite{johnson2009retrographic,zhao2023gelsight,lin2023gelsplitter}. However, extending photometric stereo from planar pads to curved fingertips introduces self-shadowing and non-uniform illumination, which can cause illumination-variation-based 3D reconstruction to fail locally. Previous studies have explored optimized illumination designs and multimodal imaging to mitigate these issues. GelSight360~\cite{tippur2023gelsight360} uses a cross-LED structure to provide sufficient illumination over an all-around curved surface, but the structure occludes part of the camera field of view. RainbowSight uses a colored LED ring to generate smooth color gradients on a semi-specular reflective layer, but it requires precise control of LED RGB intensities~\cite{tippur2024rainbowsight}. GelSplitter3D demonstrates that NIR information can improve curved-surface reconstruction, while beam-splitting and multi-camera layouts increase the system volume and require careful spatial registration across spectral images~\cite{lin20253d}. 
 
To achieve more robust photometric stereo reconstruction over the entire curved surface, this article presents a compact single-image-sensor RGB-NIR imaging design that provides an independent observation channel without introducing multi-camera registration or increasing the sensor volume.

\subsection{Vision-Based Tactile Force Estimation}
Vision-based tactile force estimation methods can be broadly divided into physics-based and data-driven approaches. Physics-based methods infer force by explicitly modeling load-induced elastomer deformation. Yang et al.~\cite{yang2023tactile} established linear fitting models between the shadow area of the loaded dome and the normal force, and between the displacement of the shadow center and the tangential force. Zhao et al.~\cite{zhao2024ifem2} discretize the gel into finite elements and use offline calibration to obtain a stiffness matrix that maps the global displacement field to the global force field. These physics-based methods provide useful interpretability, but material nonlinearity, hysteresis, boundary constraints, and complex curved contact states make accurate modeling difficult for soft fingertip sensors. Data-driven methods directly learn nonlinear force mappings using neural networks. DenseTact 2.0~\cite{do2022densetact} feeds tactile images into an encoder based on either a Swin Transformer or DenseNet, followed by two fully connected layers to regress a six-dimensional wrench. AllSight~\cite{azulay2023allsight} feeds the concatenated contact image and reference image into a modified ResNet-18, directly outputting the contact position, three-axis force, and torque around the z-axis. Nevertheless, these methods do not explicitly encode the position-dependent stiffness, thickness, curvature, and boundary constraints of curved elastomers.  

To explicitly model the position-dependent stiffness of curved elastomers, this article presents HyperForce, which combines depth-derived normal displacement and marker-derived tangential displacement and uses position-aware dynamic convolution to estimate three-axis contact force.

\subsection{Vision-Based Tactile Computing Acceleration}
Vision-based tactile perception provides rich image, geometry, force, and slip cues, but its complete processing pipeline is computationally demanding~\cite{yao2024optical}. HiVTac~\cite{quan2022hivtac} establishes a simplified thin-elastic-layer deformation model to reduce the number of markers that need to be tracked, thereby improving real-time force reconstruction capability. MinSight~\cite{andrussow2023minsight} shows that optimized camera transfer and lightweight networks can support real-time force-map estimation at 60 Hz. EveTac~\cite{funk2024evetac} uses an event camera to capture image changes, substantially reducing the amount of readout data and thereby greatly increasing the computation rate. However, these methods still depend on external CPU/GPU processing, leading to image-transfer overhead, communication latency, scheduling jitter, and higher power and integration costs, which limits deterministic low-latency feedback in compact multi-finger tactile systems. Moving tactile computation closer to the sensor is therefore important for practical dexterous-hand integration. FPGAs offer processing capability and deterministic timing, providing a suitable platform for such sensor-side computation. Oballe-Peinado et al.~\cite{oballe2017fpga} developed an FPGA-based electronics architecture for a tactile sensor suite, enabling local acquisition and preprocessing of tactile array signals. Hundhausen et al.~\cite{hundhausen2021fast} integrated five in-finger cameras with an in-hand FPGA-accelerated CNN to avoid streaming high-bandwidth raw images through the robot real-time bus, showing the benefit of decentralized image processing for reactive grasping.

Nevertheless, existing edge tactile systems have not yet demonstrated an end-to-end deployment of the complete vision-based tactile processing pipeline. In contrast, our work implements the full pipeline, including image-stream processing, 3D reconstruction, and normal force estimation, on an FPGA-based edge platform, thereby reducing raw-data transfer and enabling deterministic low-latency feedback for compact dexterous-hand integration.

\section{FasTac Sensor Design}

\begin{figure}[t]
  \centering
  \includegraphics[width=\linewidth]{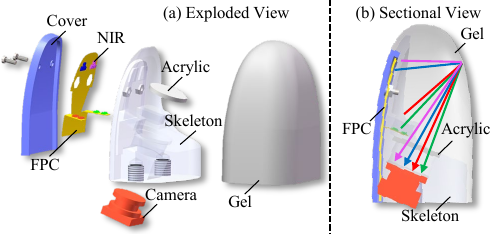}
  \caption{Mechanical design and integration of FasTac.}
  \label{fig:sensor_structure}
\end{figure}
\subsection{Overall Structure}
The mechanical design of FasTac prioritizes compactness and integration for dexterous manipulation. As shown in Fig.~\ref{fig:sensor_structure}, the sensor mainly consists of three modules: a light-guiding fingertip skeleton that houses the camera, a multispectral LED FPC secured by the rear cover, and a sensing interface formed by the gel skin.

\subsection{Imaging Module}
FasTac adopts a custom-designed miniature camera module. The camera module is controlled by an onboard FPGA, which receives raw data through the MIPI interface. The image sensor is an OmniVision OV2736 RGB-IR CMOS sensor, which uses a unique 4x4 RGB-IR color filter array (CFA) to simultaneously capture visible and near-infrared information in a single exposure. Compared with prism-based multispectral schemes\cite{lin20253d}, the RGB and NIR signals of this module originate from the same optical center, enabling intrinsic pixel-level alignment and smaller volume.

\subsection{Illumination System}
Building on the conventional RGB configuration, this work introduces an NIR light source as a fourth independent channel. This design aims to satisfy the three-source simultaneous illumination assumption of photometric stereo on curved geometries using a simple illumination structure, thereby ensuring the robustness of curved-surface 3D reconstruction.

To improve maintainability and integration, we designed a Flexible Printed Circuit (FPC) that conforms to the dorsal curvature of the fingertip skeleton. LED arrays in four colors (R, G, B, and NIR) are placed at the four corners of the FPC, with the illumination directions of the red/green sources perpendicular to those of the blue/NIR sources. This arrangement increases the angular diversity and spatial distinctiveness of the illumination vectors, ensuring the numerical stability of the photometric solution. Furthermore, placing the light sources behind the skeleton exploits the translucency and scattering properties of the photosensitive-resin skeleton, transforming discrete LED point sources into a soft and uniform diffuse light field. This reduces specular highlights and improves illumination uniformity.

\subsection{Tactile Skin}
We use platinum-catalyzed silicone rubber (Smooth-On Solaris) as the matrix material and mix Solaris A, Solaris B, and Slacker softener at a weight ratio of 1:1:3. The high proportion of Slacker gives the elastomer excellent compliance, enabling it to sensitively capture minute contact deformations. After vacuum degassing, the mixture is poured into a mold pre-assembled with the fingertip skeleton. To eliminate surface artifacts caused by 3D-printing layer striations, the bottom mold is made from precision-machined optical-grade acrylic (PMMA). After curing at room temperature for 24 hours, the elastomer forms a robust integrated structure with the skeleton.

To form an ideal optical reflective layer on the elastomer surface, we formulated a specialized reflective paint. The coating consists of 800-mesh silver powder, silicone solvent, and thinner, and is uniformly applied to the elastomer surface using an airbrush. The silicone solvent ensures strong adhesion between the coating and the substrate; the silver powder particles block external light while improving image contrast; and the thinner controls the flowability and smoothness of the paint, ensuring the uniformity and integrity of the coating.

After the reflective layer is coated and cured, a laser marking process is used to ablate a marker-dot array in the sensing region on the gel surface. Black silicone ink is then applied to the marker-dot regions, so that the markers appear black in the captured images.

\section{FasTac 3D Shape Reconstruction and Force Estimation Pipeline}
\begin{figure*}[!t]
  \centering
  \includegraphics[width=\textwidth]{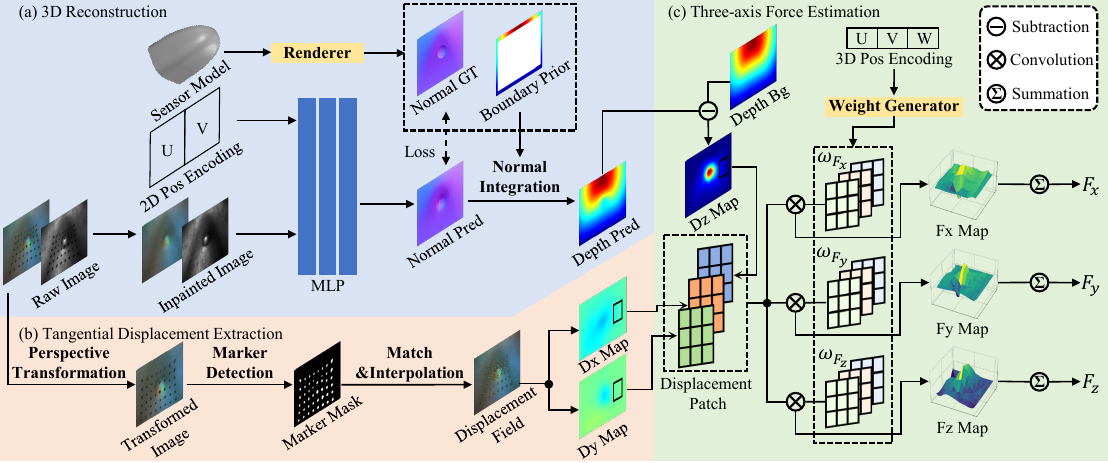}
  \caption{3D reconstruction, Tangential displacement extraction, and three-axis force estimation pipeline.}
  \label{fig:pipeline}
\end{figure*}

The overall perception pipeline is illustrated in Fig.~\ref{fig:pipeline}, including multispectral 3D reconstruction, marker-based tangential displacement extraction, 3D displacement fusion, and HyperForce-based three-axis force estimation.

\subsection{Multispectral Photometric Stereo for 3D Shape Reconstruction}
\subsubsection{Four-Source Photometric Stereo}
The geometric perception of FasTac is based on photometric stereo~\cite{woodham1979photometric}. For a Lambertian surface with normal $\boldsymbol{n}$ and albedo $\rho$, the observed intensity under illumination direction $\boldsymbol{l}$ is $I=\rho\boldsymbol{l}^{\mathrm{T}}\boldsymbol{n}$. A standard RGB three-source configuration usually estimates the surface normal from $\mathbf{L}_{obs}=[\boldsymbol{l}_{\mathrm{R}},\boldsymbol{l}_{\mathrm{G}},\boldsymbol{l}_{\mathrm{B}}]^{\mathrm{T}}$. However, in a compact curved fingertip structure, RGB light sources cannot cover the entire surface, which may lead to $\mathrm{rank}(\mathbf{L}_{obs})<3$ and therefore cause underdetermined surface normals and blurred reconstruction. To achieve robust surface normal estimation with a simple illumination structure, FasTac introduces NIR as the fourth independent spectral channel and constructs $\mathbf{L}=[\boldsymbol{l}_{\mathrm{R}},\boldsymbol{l}_{\mathrm{G}},\boldsymbol{l}_{\mathrm{B}},\boldsymbol{l}_{\mathrm{NIR}}]^{\mathrm{T}}\in\mathbb{R}^{4\times3}$, forming an over-determined system $\boldsymbol{I}=\rho\mathbf{L}\boldsymbol{n}$. Thus, even if deformation occludes one channel, the remaining three channels can still form a valid basis, improving surface normal estimation stability on curved surfaces.

\begin{figure}[t]
  \centering
  \includegraphics[width=\linewidth]{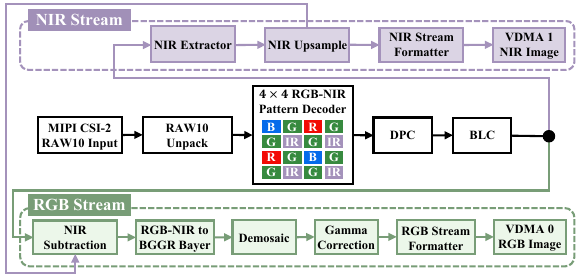}
  \caption{RGB-NIR spectral demultiplexing and preprocessing pipeline. The single-channel RAW10 stream is decoded from the RGB-IR CFA into pixel-aligned RGB and NIR images.}
  \label{fig:demultiplexing}
\end{figure}

\subsubsection{Spectral Demultiplexing and Pre-processing}
As shown in Fig.~\ref{fig:demultiplexing}, the RAW10 data produced by the OV2736 RGB-IR CMOS sensor are first unpacked from the MIPI CSI-2 stream, yielding $I_{\rm raw}(u,v)\in[0,2^{10}-1]$. The sensor adopts a $4\times4$ RGB-IR CFA pattern, in which R, G, B, and IR pixels are classified by the phase $\phi(u,v)=(v\bmod4,u\bmod4)$. The preprocessing flow sequentially performs dead-pixel correction (DPC), black-level correction (BLC), NIR sampling and upsampling, IR crosstalk subtraction, standard BGGR Bayer reconstruction, demosaicing, and output mapping. The NIR channel is sampled from odd rows and odd columns as $I^s_{\mathrm{NIR}}(m,n)=I_{\rm blc}(2m+1,2n+1)$ and then interpolated into a dense image $I_{\mathrm{NIR}}$. This NIR estimate is subsequently used for IR subtraction of the visible pixels:
\begin{equation}
\begin{aligned}
\tilde I(u,v)=\operatorname{clip}\Bigl(
&I_{\rm blc}(u,v)\\
&-\alpha\min\!\left\{
I_{\rm NIR}(u,v),I_{\max}/\beta
\right\},\,0,I_{\max}\Bigr).
\end{aligned}
\end{equation}
where $I_{\rm blc}$ is the black-level-corrected raw image, $\tilde{I}(u,v)$ is the IR-corrected visible-channel intensity, and $\operatorname{clip}(\cdot,a,b)$ truncates its input to the interval $[a,b]$. $\alpha$ controls the IR subtraction strength, $\beta$ prevents overcorrection by limiting the NIR estimate, and $I_{\max}$ is the RAW10 saturation value. The corrected RGB-IR CFA is rearranged into a standard BGGR Bayer image and demosaiced to obtain an RGB image; the NIR image is mapped to an 8-bit grayscale image according to the RAW10 dynamic range. The final output is $\boldsymbol{I}(u,v)=[I_{\mathrm{B}},I_{\mathrm{G}},I_{\mathrm{R}},I_{\mathrm{NIR}}]^{\mathrm{T}}$. Since RGB and NIR are captured by the same sensor, under the same exposure, and through the same optical center, this flow avoids the registration errors of multi-camera or beam-splitting multispectral structures~\cite{lin2023gelsplitter,lin20253d}, providing synchronized and zero-parallax multispectral inputs for curved tactile reconstruction.

\subsubsection{Position-Aware MLP for Surface Normal Estimation}
\label{sec:mlp}
To achieve pixel-wise surface normal estimation on the curved fingertip, we adopt a physics-prior guided lightweight MLP. Because the internal space of FasTac is compact, the LEDs exhibit near-field characteristics, and the sensing surface is non-planar, the illumination vector is no longer a global constant but varies with pixel position $\boldsymbol{p}=(u,v)$ as $\mathbf{L}(\boldsymbol{p})$. Therefore, the light intensity satisfies $\boldsymbol{I}_{u,v}=\mathcal{R}(\boldsymbol{n}_{u,v},\mathbf{L}(\boldsymbol{p}),\rho)$, where $\mathcal{R}(\cdot)$ denotes the local image-formation function. If only intensity is used as input, the network struggles to distinguish shadows caused by surface inclination from brightness variations caused by distance to the light sources. We concatenate normalized coordinates as positional encoding with the four-channel spectral intensity and learn $\hat{\boldsymbol{n}}_{u,v}\approx\mathcal{F}(\boldsymbol{I}_{u,v}\oplus\boldsymbol{p}_{u,v})$, where $\oplus$ denotes feature concatenation. Since this mapping is mainly determined by the single-point intensity and position rather than neighborhood texture, we use a pixel-wise MLP with a 6-dimensional input vector, and a final output $\hat{\boldsymbol{n}}=[n_x,n_y,n_z]$. Before computing the training loss, both the predicted and ground-truth surface normals are normalized to unit length. The surface normal estimation network is optimized using an equally weighted component-wise L1 loss:
\begin{equation}
\mathcal{L}_{n}
=
\frac{1}{|\Omega|}
\sum_{\boldsymbol{p}\in\Omega}
\sum_{\alpha\in\{x,y,z\}}
\left|
\hat n_{\alpha}(\boldsymbol{p})
-
n^{*}_{\alpha}(\boldsymbol{p})
\right|,
\end{equation}
where $\Omega$ is the pixel domain, $\alpha\in\{x,y,z\}$ indexes the surface normal component, and $\hat{\boldsymbol{n}}$ and $\boldsymbol{n}^{*}$ are the unit-normalized predicted and ground-truth surface normals, respectively.

\subsubsection{Fast Poisson Depth Reconstruction with Boundary Prior}
\label{sec:boundary_prior_depth}
After surface normal estimation, FasTac reconstructs contact depth by integrating the estimated surface normals. Standard fast Poisson solvers usually imply the zero-boundary condition, but the boundary of a curved fingertip has non-zero depth related to the geometry. We therefore use a Dirichlet-constrained fast Poisson solver and embed the boundary prior $D_{\text{prior}}$ derived from the sensor CAD model into the source term:
\begin{equation}
\tilde f_{i,j}=f_{i,j}-\sum_{(m,n)\in\mathcal{N}(i,j)\cap\partial\Omega}D_{\text{prior}}(m,n).
\end{equation}
Here, $f_{i,j}$ is the original Poisson source term at pixel $(i,j)$, $\tilde f_{i,j}$ is the boundary-corrected source term, $\mathcal{N}(i,j)$ denotes the four-connected neighborhood, and $\partial\Omega$ denotes the boundary of the valid reconstruction domain. This operation moves the known boundary values to the right-hand side of the discrete Laplacian system, so the standard type-I discrete sine transform (DST-I) can still be used to solve the curved-surface depth efficiently while avoiding the depth drift caused by the zero-boundary assumption.

\subsection{FEM-inspired Dynamic Convolution for Three-Axis Force Estimation}

Variations in thickness, curvature, and boundary constraints induce position-dependent effective stiffness in a curved elastomer, so the same displacement may correspond to different mechanical mappings in different regions. To describe this spatial nonuniformity, we construct a 3D displacement field from reconstructed depth and marker motion, and then map it to three-axis contact force using HyperForce, an FEM-inspired dynamic convolution network.

\subsubsection{3D Displacement Field Extraction via Depth Reconstruction and Marker Tracking}
\label{sec:dis_extraction}
The 3D displacement field is defined as $\boldsymbol{u}(\boldsymbol{p})=[d_x(\boldsymbol{p}),d_y(\boldsymbol{p}),d_z(\boldsymbol{p})]^{\mathrm{T}}$. Tangential displacement is obtained by tracking embedded markers: sparse marker displacements $\boldsymbol{d}_k$ are first extracted from $M$ markers using thresholding and blob detection, and a radial basis function (RBF) interpolator is then used to recover the dense tangential field $\boldsymbol{u}_{\tan}(\boldsymbol{p})=[d_x(\boldsymbol{p}),d_y(\boldsymbol{p})]^{\mathrm{T}}$. We use the inverse multiquadric (IMQ) kernel
\begin{equation}
\boldsymbol{u}_{\tan}(\boldsymbol{p})=\sum_{k=1}^{M}\boldsymbol{w}_k\phi(|\boldsymbol{p}-\boldsymbol{c}_k|),\qquad
\phi(r)=\frac{1}{\sqrt{r^2+\varepsilon^2}},
\end{equation}
where $\boldsymbol{c}_k$ is the detected marker center, $\boldsymbol{w}_k$ is the RBF weight solved from sparse marker constraints, $r=|\boldsymbol{p}-\boldsymbol{c}_k|$ is the pixel-domain distance, and $\varepsilon$ is the kernel shape parameter. The kernel decay form is consistent with the deformation response of a semi-infinite elastic body under point loading. The normal displacement $d_z(\boldsymbol{p})$ is obtained from the difference between the background depth $D_0(\boldsymbol{p})$ and the current reconstructed depth $D_t(\boldsymbol{p})$: 
\begin{equation}
d_z(\boldsymbol{p})=D_0(\boldsymbol{p})-D_t(\boldsymbol{p}).
\end{equation}

\subsubsection{FEM-inspired Dynamic Convolution Network}
\label{sec:Hyperforce}

HyperForce is motivated by finite element analysis (FEM) for vision-based tactile force reconstruction~\cite{ma2019dense,zhao2024ifem2}. In FEM, the elastomer is discretized into finite elements whose vertices are referred to as nodes. Given the displacements of all nodes, assembled into the global displacement vector $\boldsymbol{U}$, the corresponding nodal contact force vector $\boldsymbol{F}$ can be estimated using the global stiffness matrix $\mathbf{K}$:
\begin{equation}
\boldsymbol{F}=\mathbf{K}\boldsymbol{U}.
\end{equation}

Since $\mathbf{K}$ is sparse, the force at a node is mainly determined by the displacements in its local neighborhood. Therefore, by treating each pixel as a node, the local multiplication in FEM can be approximated by a convolution-like operation:
\begin{equation}
\boldsymbol{f}(\boldsymbol{p})\approx
\sum_{\boldsymbol{q}\in\mathcal{N}}
\mathbf{G}_{\boldsymbol{p},\boldsymbol{q}}
\boldsymbol{u}(\boldsymbol{p}+\boldsymbol{q}),
\end{equation}
where $\boldsymbol{p}$ is the current pixel or node, $\boldsymbol{q}$ is a relative offset in the neighborhood $\mathcal{N}$, $\boldsymbol{u}(\boldsymbol{p}+\boldsymbol{q})=[d_x,d_y,d_z]^{\mathrm{T}}$ is the local 3D displacement, $\boldsymbol{f}(\boldsymbol{p})=[f_x,f_y,f_z]^{\mathrm{T}}$ is the local 3D force response, and $\mathbf{G}_{\boldsymbol{p},\boldsymbol{q}}$ is the local stiffness-like coupling weight.

For a homogeneous planar elastomer with a regular mesh, this coupling weight can be approximated as $\mathbf{G}_{\boldsymbol{q}}$ because the local stiffness pattern is nearly translation invariant, so FEM can be represented by a globally shared convolution kernel. In contrast, for the curved FasTac elastomer, $\mathbf{G}_{\boldsymbol{p},\boldsymbol{q}}$ must depend on $\boldsymbol{p}$ because local normal directions, curvature, effective thickness, and boundary constraints vary across the surface. HyperForce therefore approximates the curved-surface FEM operator using position-dependent dynamic convolution. For each $\boldsymbol{p}$, the $K\times K$ displacement patch is vectorized as $\boldsymbol{v}(\boldsymbol{p})\in\mathbb{R}^{3K^2}$. A lightweight hypernetwork implemented by $1\times1$ convolutional layers generates the dynamic kernel from the normalized 3D coordinate encoding $\boldsymbol{c}(\boldsymbol{p})$:
\begin{equation}
\mathbf{w}_{\alpha}(\boldsymbol{p})
=
\mathcal{H}_{\alpha}(\boldsymbol{c}(\boldsymbol{p})),
\quad
\mathbf{w}_{\alpha}(\boldsymbol{p})\in\mathbb{R}^{3K^2},
\quad
\alpha\in\{x,y,z\}.
\end{equation}
Here, $\mathcal{H}_{\alpha}$ is the coordinate-conditioned weight generator for force component $\alpha$, and $\mathbf{w}_{\alpha}(\boldsymbol{p})$ is the dynamic kernel applied at $\boldsymbol{p}$. The pixel-wise force component is computed as
\begin{equation}
\hat{f}_{\alpha}(\boldsymbol{p})
=
\mathbf{w}_{\alpha}(\boldsymbol{p})^{\mathrm{T}}
\boldsymbol{v}(\boldsymbol{p}).
\end{equation}
Finally, the resultant three-axis force is obtained over the valid tactile region $\Omega$:
\begin{equation}
\hat{\boldsymbol{F}}
=
\sum_{\boldsymbol{p}\in\Omega}
\hat{\boldsymbol{f}}(\boldsymbol{p}).
\end{equation}

HyperForce is trained using global three-axis force supervision. After spatially integrating the pixel-wise intermediate responses, the predicted force is compared with the force-sensor reference using an equally weighted L1 loss:
\begin{equation}
\mathcal{L}_{F}
=
\sum_{\alpha\in\{x,y,z\}}
\left|
\hat F_{\alpha}
-
F^{*}_{\alpha}
\right|,
\end{equation}
where $\hat F_{\alpha}$ and $F^{*}_{\alpha}$ are the predicted and reference forces along axis $\alpha$, respectively.

\subsection{FPGA-Based Edge Deployment of the Image-to-$F_z$ Estimation Pipeline}
\begin{figure*}[!t]
  \centering
  \includegraphics[width=\textwidth]{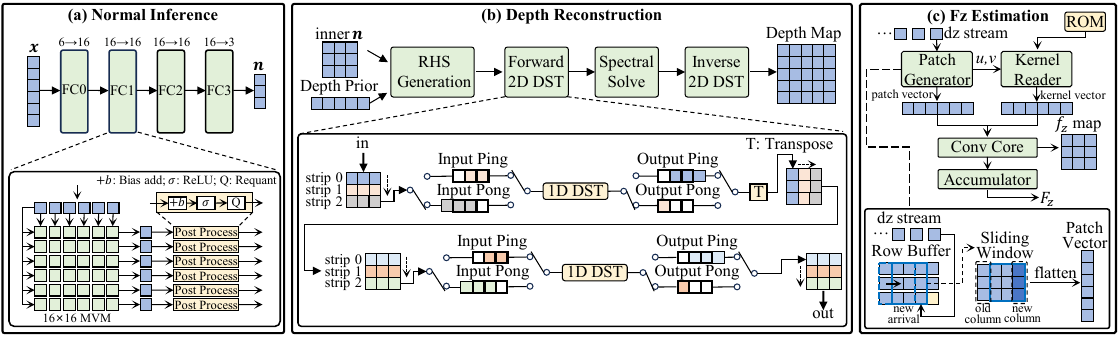}
  \caption{\textbf{FPGA architecture of the image-to-$F_z$ pipeline.} (a) Quantized surface normal estimation. (b) Boundary-prior depth reconstruction. (c) Streaming $F_z$ estimation.}
  \label{fig:fpga_architecture}
\end{figure*}

To meet dynamic-feedback requirements, we deploy the normal-force branch of FasTac on an FPGA edge platform. The system adopts a heterogeneous architecture with a processing system (PS) and programmable logic (PL) to execute the image-to-$F_z$ pipeline from surface normal estimation to boundary-prior depth reconstruction and position-aware $F_z$ estimation. The PS handles frame-level orchestration, including image synchronization, background and metadata management, Advanced eXtensible Interface Lite (AXI-Lite) control, and result readout, while the PL implements the latency-critical streaming datapath with internal FIFOs between the three computation modules. Inputs are cropped and downsampled to accommodate the limited FPGA buffering and computation resources and to increase the camera frame rate through a smaller sensor readout window.

\subsubsection{Surface Normal Estimation Module}
The surface normal estimation module maps the input feature
$\boldsymbol{x}(\boldsymbol{p})=[\Delta I_{\mathrm{R}},\Delta I_{\mathrm{G}},\Delta I_{\mathrm{B}},\Delta I_{\mathrm{NIR}},p_x,p_y]^{\mathrm{T}}$
to the pixel-wise normal
$\hat{\boldsymbol{n}}(\boldsymbol{p})=[\hat n_x,\hat n_y,\hat n_z]^{\mathrm{T}}$.
Here, $\boldsymbol{p}$ is the pixel coordinate, the $\Delta I$ terms are intensity differences, and $p_x,p_y$ are normalized coordinates. The deployed model remains a lightweight MLP, as shown in Fig.~\ref{fig:fpga_architecture}(a). Each FC layer is a fully unrolled $16\times16$ matrix-vector multiplication (MVM) tile: 16 input lanes and 16 output-neuron lanes form up to 16 output dot products concurrently. Products are reduced as grouped partial sums and then accumulated, shortening adder chains for timing closure. For FC0--FC2, the biased MVM output $a_i^{(l)}$ passes through ReLU and requantization,
\begin{equation}
q_i^{(l+1)}=
\operatorname{clip}_{[0,127]}\!\left(
\left\lfloor \frac{M_l}{2^{r_l}}\max(0,a_i^{(l)})\right\rfloor
\right),\quad l=0,1,2 ,
\end{equation}
where $q_i^{(l+1)}$ is the next-layer activation of neuron $i$, $a_i^{(l)}$ is the biased MVM output of layer $l$, $M_l$ and $r_l$ are the integer multiplier and shift, \(\lfloor\cdot\rfloor\) denotes the floor operation, while $\operatorname{clip}_{[0,127]}(\cdot)$ saturates to the unsigned activation range. FC3 adds bias only, without ReLU or requantization. Unlike the hidden-layer activations, its three signed outputs are not consumed by another quantized MVM; instead, they are retained in the wider accumulator format and passed directly to the gradient ratios $p=-\hat n_x/\hat n_z$ and $q=-\hat n_y/\hat n_z$. The module processes the pixels as a continuous stream: it neither waits for a full frame nor restarts a full control sequence per pixel.
\subsubsection{Depth Reconstruction Module}
The depth-reconstruction module converts estimated surface normals to gradients $p$ and $q$, and solves the boundary-prior Poisson problem over the inner $126\times126$ region with a one-pixel boundary width. The right-hand side (RHS) generation module computes divergence and injects the Dirichlet boundary prior:
\begin{equation}
\begin{aligned}
b_{i,j}={}&(p_{i,j}-p_{i,j-1})+(q_{i,j}-q_{i-1,j})\\
&-\sum_{(m,n)\in\mathcal{N}(i,j)\cap\partial\Omega}
D_{\mathrm{prior}}(m,n),
\end{aligned}
\end{equation}
where $b_{i,j}$ is the RHS value at inner pixel $(i,j)$, $\mathcal{N}(i,j)$ is its discrete neighbor set, $\partial\Omega$ is the boundary, and $D_{\mathrm{prior}}(m,n)$ is the known boundary depth. The boundary-corrected RHS is then processed by an orthonormal DST-I Poisson solver consisting of a separable forward 2-D DST, element-wise division by the Poisson eigenvalues, and a separable inverse 2-D DST. As shown in the lower part of Fig.~\ref{fig:fpga_architecture}(b), each 2-D transform is executed as a row transform, a matrix transpose, and a second row transform. The matrix is partitioned into strips, and ping-pong buffers overlap the write, compute, and read phases of adjacent strips. This organization hides memory movement behind the 1-D DST engine and improves streaming throughput, while independent forward and inverse executors allow the two transform phases to be scheduled separately.

\subsubsection{Normal-Force Estimation Module}
We deploy only the normal-force branch because $F_z$ is the most direct and highest-frequency feedback variable for contact detection, preload control and grasp stability. When contact is mainly normal, tangential displacement is small compared with normal displacement, so $F_z$ can be inferred from normal displacement with a regular stream suitable for FPGA pipelining.
The normal-force estimation module first computes
\begin{equation}
q_{\Delta z}(\boldsymbol{p})=Q_z\!\left(\max(Z_0(\boldsymbol{p})-\hat Z_t(\boldsymbol{p}),0)\right),
\end{equation}
where $q_{\Delta z}$ is the quantized normal displacement, $Q_z(\cdot)$ is displacement quantization, $Z_0$ is the background depth, $\hat Z_t$ is the current depth, and $\boldsymbol{p}$ is the pixel coordinate. The hardware preserves the position-aware mapping in HyperForce. For each pixel, the patch generator emits a $K\times K$ vector while the kernel reader simultaneously returns a coordinate-indexed vector from Read-Only Memory (ROM):
\begin{equation}
\boldsymbol{v}_{\boldsymbol{p}}=
\{q_{\Delta z}(\boldsymbol{p}+\boldsymbol{\delta}_m)\}_{m=1}^{K^2},
\qquad
\boldsymbol{k}_{\boldsymbol{p}}=\operatorname{ROM}(u,v),
\end{equation}
where $K$ is the side length of the local displacement window, $\boldsymbol{\delta}_m$ indexes the $m$-th window offset, $\boldsymbol{v}_{\boldsymbol{p}}$ is the displacement patch vector, $\boldsymbol{k}_{\boldsymbol{p}}$ is the position-aware kernel vector, and $(u,v)$ is the pixel coordinate used for ROM addressing. The deployed force computation uses
\begin{equation}
\begin{aligned}
\hat f_z(\boldsymbol{p})&=\sum_{m=1}^{K^2}
k_{\boldsymbol{p},m}v_{\boldsymbol{p},m},\\
A_z&=\sum_{\boldsymbol{p}\in\Omega}\hat f_z(\boldsymbol{p}),\\
\hat F_z&=s_F A_z,\qquad s_F=\frac{1}{s_ks_d}.
\end{aligned}
\end{equation}
Here, $\hat f_z(\boldsymbol{p})$ is the pixel-wise integer response, $A_z$ is the raw full-frame accumulator, and $s_k$ and $s_d$ are the quantization scales for the kernel and normal displacement, respectively. The FPGA first performs convolution and full-frame accumulation using the quantized kernel and displacement. Since the integer computation scales the corresponding floating-point result by \(s_ks_d\), \(s_F=(s_ks_d)^{-1}\) converts the raw accumulator to the force estimate in Newtons. The patch generator stores only the recent $K$ rows. Away from the first column, the window shifts left and fetches the new rightmost column; at a row start, the whole window is refreshed from row buffers. A patch is emitted when the last required sample arrives, avoiding full-frame caching.

\section{Experiments}

\subsection{Data Collection and Training Protocol}
\begin{figure}[t]
  \centering
  \includegraphics[width=\linewidth]{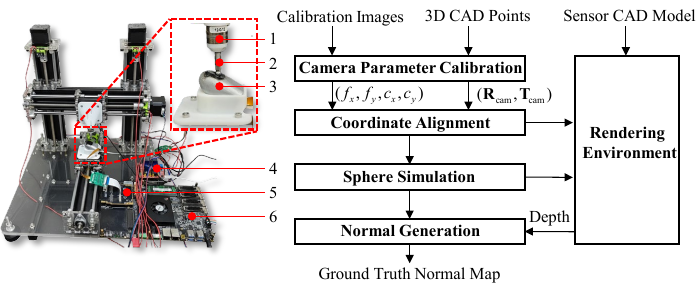}
  \caption{Data acquisition and geometric ground-truth generation pipeline. Labels: 1, ATI Nano17 force/torque sensor; 2, spherical indenter; 3, FasTac; 4, ESP32 CNC controller; 5, FPGA Mezzanine Card (FMC)-to-MIPI adapter; and 6, FPGA development board.}
  \label{fig:data_collection}
\end{figure}

Fig.~\ref{fig:data_collection} summarizes the automated acquisition pipeline. A force-sensor-mounted spherical indenter applied controlled normal and tangential contacts while tactile images, CNC poses, and force readings were recorded synchronously. Pixel-wise normal and depth ground truth was rendered from the pose, indenter geometry, and aligned sensor CAD model. Spatially disjoint datasets were used for training, validation, and testing. The complete acquisition protocol, dataset composition, and training hyperparameters are provided in Section~I of the supplementary material.

\begin{figure*}[!t]
  \centering
  \includegraphics[width=\textwidth]{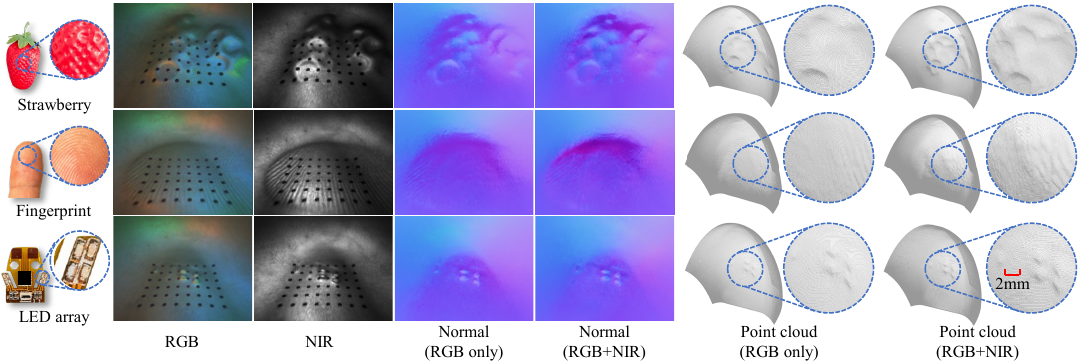}
    \caption{%
    \textbf{3D reconstruction results on three representative objects.}
    The RGB/NIR inputs, predicted normal maps, and reconstructed point clouds under RGB-only and RGB-NIR illumination are compared.
    }
  \label{fig:3d_reconstruction}
\end{figure*}

\begin{figure}[!t]
  \centering
  \includegraphics[width=\linewidth]{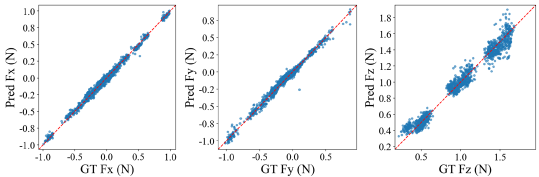}
  \caption{\textbf{Regression between predicted three-axis contact forces and ATI ground truth.}}
  \label{fig:force_pred}
\end{figure}

\begin{table}[t]
\centering
\caption{Gradient-error comparison between RGB-only and RGB-NIR illumination.}
\label{tab:reconstruction_accuracy}
\renewcommand{\arraystretch}{1.2}
\footnotesize
\begin{tabular*}{\columnwidth}{c @{\extracolsep{\fill}} ccc}
\hline
\multirow{2}{*}{Component} & \multicolumn{2}{c}{Gradient Error (mm/pixel)} & \multirow{2}{*}{\makecell{Improv.}} \\ \cline{2-3}
 & RGB-only & RGB-NIR &  \\ 
\hline
$G_x$ (MAE) & 0.0047 & \textbf{0.0028} & $\downarrow$ 40\% \\
$G_y$ (MAE) & 0.0051 & \textbf{0.0040} & $\downarrow$ 22\% \\
Total (MAE) & 0.0098 & \textbf{0.0068} & $\downarrow$ 31\% \\
\hline
\end{tabular*}
\end{table}

\begin{table}[t]
\centering
\caption{Depth MAE under different illumination and boundary conditions.}
\label{tab:depth_reconstruction_accuracy}
\renewcommand{\arraystretch}{1.2}
\footnotesize
\begin{tabular*}{\columnwidth}{c @{\extracolsep{\fill}} cccc}
\hline
\multirow{2}{*}{Metric (MAE / mm)} & \multicolumn{2}{c}{RGB-only} & \multicolumn{2}{c}{RGB-NIR} \\ \cline{2-5}
 & \makecell{w/ depth\\prior} & \makecell{w/o depth\\prior} & \makecell{w/ depth\\prior} & \makecell{w/o depth\\prior} \\
\hline
\makecell{Depth error} & 0.0618 & 0.2730 & \textbf{0.0415} & 0.2686 \\
\hline
\end{tabular*}
\end{table}

\subsection{3D Geometry Reconstruction Evaluation}
To evaluate fine geometric perception required in dexterous manipulation, we first evaluate surface normal estimation and depth reconstruction on a spherical-indenter dataset with radius $r=2.5{\text{mm}}$. As shown in Table~\ref{tab:reconstruction_accuracy}, RGB-NIR reduces the gradient MAE of $G_x$ and $G_y$ by $40\%$ and $22\%$, respectively, and reduces the total gradient MAE from 0.0098 to 0.0068 mm/pixel. This confirms that the fourth NIR observation provides additional constraints for surface normal estimation on curved surfaces.

Depth is evaluated by the MAE between the reconstructed depth map and the CAD-aligned ground truth. In Table~\ref{tab:depth_reconstruction_accuracy}, RGB-NIR with the depth prior achieves 0.0415 mm MAE, compared with 0.0618 mm for RGB-only, corresponding to a $32.8\%$ reduction. Without the depth prior, both illumination schemes drift to about 0.27 mm MAE, showing that the boundary-prior constraint is essential for stable curved-surface reconstruction.

We further evaluate real-object reconstruction on a strawberry surface, a fingerprint, and an LED array. As shown in Fig.~\ref{fig:3d_reconstruction}, compared with RGB-only reconstruction, RGB-NIR produces more complete surface pits, better preserves ridge-valley continuity, and resolves miniature LED pads more clearly.

\subsection{Three-Axis Force Sensing Evaluation}
\begin{table}[t]
\centering
\caption{Ablation of HyperForce force estimation modes.}
\label{tab:force_ablation}
\footnotesize
\renewcommand{\arraystretch}{1.08}
\begin{tabular*}{\columnwidth}{@{\extracolsep{\fill}}>{\centering\arraybackslash}m{0.16\columnwidth}>{\centering\arraybackslash}m{0.25\columnwidth}ccc@{}}
\hline
\multirow[c]{2}{*}{Kernel} & \multirow[c]{2}{*}{Input} & \multicolumn{3}{c}{Component MAE (N)} \\
\cline{3-5}
 & & $F_x$ & $F_y$ & $F_z$ \\
\hline
\multirow[c]{3}{*}{Dynamic} & $d_x+d_y$ & 0.0386 & 0.0279 & 0.1570 \\
 & $d_z$ & 0.0235 & 0.0325 & 0.0589 \\
 & $d_x+d_y+d_z$ & \textbf{0.0235} & \textbf{0.0246} & \textbf{0.0545} \\
\hline
\multirow[c]{3}{*}{Fixed} & $d_x+d_y$ & 0.0545 & 0.0318 & 0.5123 \\
 & $d_z$ & 0.1028 & 0.0637 & 0.4980 \\
 & $d_x+d_y+d_z$ & 0.0516 & 0.0344 & 0.4905 \\
\hline
\end{tabular*}
\end{table}

We use a six-axis ATI sensor as the reference for three-axis force evaluation. Mean absolute error (MAE) and root-mean-square error (RMSE) are computed over the test set. Their normalized forms, normalized mean absolute error (NMAE) and normalized root-mean-square error (NRMSE), are obtained by normalization with the maximum ground-truth load of each force component. Fig.~\ref{fig:force_pred} shows strong linear agreement between predicted and reference forces. The MAE values are 0.0235, 0.0246, and 0.0545 N for $F_x$, $F_y$, and $F_z$, corresponding to NMAE values of 2.37\%, 2.41\%, and 2.74\%; the NRMSE values are 3.15\%, 3.33\%, and 3.72\%, with $R^2$ values of 0.9921, 0.9873, and 0.9690.

Table~\ref{tab:force_ablation} verifies the effect of the dynamic kernel and 3D displacement input. For the same input, the position-aware dynamic kernel consistently outperforms the fixed CNN, especially in $F_z$: with full $d_x+d_y+d_z$ input, $F_z$ MAE decreases from 0.4905 N to 0.0545 N. Under the dynamic setting, full 3D displacement gives the best overall accuracy, indicating that normal indentation and tangential marker displacement provide complementary force cues.

\begin{figure}[t]
  \centering
  \includegraphics[width=\linewidth]{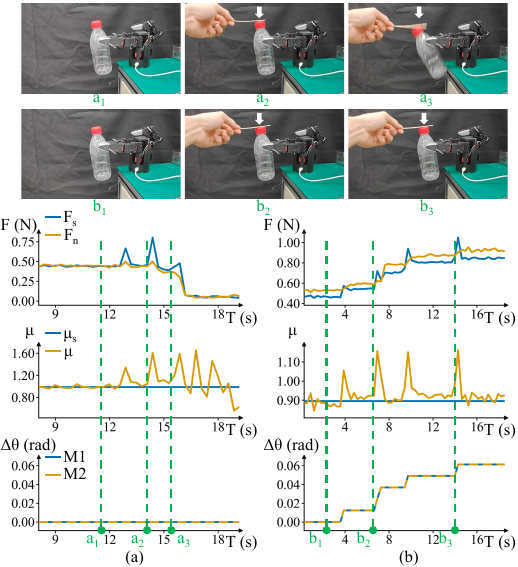}
  \caption{\textbf{Contact force $F$, friction coefficient $\mu$, and motor angle variation $\Delta\theta$ without and with friction-coefficient feedback.} Panel (a) shows the case without feedback, while panel (b) shows the case with feedback. The green dashed lines indicate representative time points during the experiment.}
  \label{fig:force_feedback_grasp_result}
\end{figure}

\begin{figure*}[t]
  \centering
  \includegraphics[width=\textwidth]{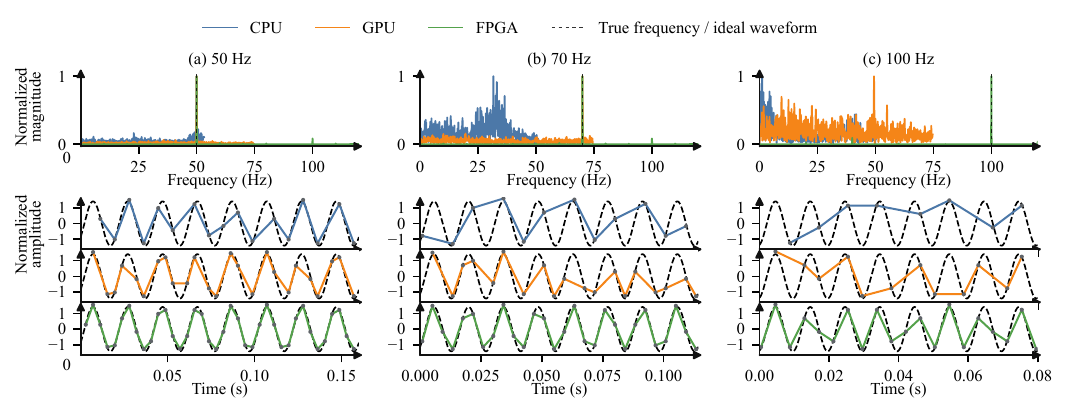}
  \caption{\textbf{Frequency spectra (top row) and time-domain normal-force responses (bottom row) of CPU, GPU, and FPGA under 50 Hz, 70 Hz, and 100 Hz inputs.}}
  \label{fig:vibration_frequency_response}
\end{figure*}

To validate stable grasp feedback under external disturbances, we conduct a friction-coefficient-based grasping task. As shown in Fig.~\ref{fig:force_feedback_grasp_result}, a dexterous hand grasps a plastic bottle with two FasTac fingertips while repeated downward disturbances are applied to the bottle cap. The controller computes
\begin{equation}
F_s=\sqrt{F_x^2+F_y^2},\qquad F_n=F_z,\qquad \mu=\frac{F_s}{F_n}.
\end{equation}
Here, $\mu_s$ and $d$ are the experimentally determined stable friction coefficient and safety margin, respectively. When $\mu>\mu_s+d$, the inter-finger angle of the two distal joints is reduced to increase grasp stability. Fig.~\ref{fig:force_feedback_grasp_result} compares the contact-force, friction-coefficient, and motor-angle responses without and with feedback. Without feedback, repeated disturbances drive $F_s$ and $\mu$ upward while the motor angle remains unchanged, and the subsequent force drop indicates slip, as shown at point $\mathrm{a}_3$. With feedback enabled, each threshold crossing triggers grasp tightening, increasing $F_n$ and returning $\mu$ to the stable region. This demonstrates real-time closed-loop use of FasTac three-axis force estimation.

\subsection{Edge Deployment and High-Speed Tactile Processing Evaluation}

\begin{table}[t]
\centering
\caption{End-to-end image-to-normal-force performance of CPU, GPU, and FPGA implementations.}
\label{tab:edge_deployment_performance}
\scriptsize
\setlength{\tabcolsep}{3pt}
\renewcommand{\arraystretch}{1.15}
\resizebox{\columnwidth}{!}{%
\begin{tabular}{@{} c c c c @{}}
\hline
\textbf{Platform} & \textbf{Latency (ms)} & \textbf{Force MAE (N)} & \textbf{Energy (mJ/frame)} \\
\hline
CPU & 6.82 & 0.0669 & 238.05 \\
GPU & 3.26 & \textbf{0.0667} & 33.60 \\
FPGA & \textbf{1.09} & 0.0679 & \textbf{8.41} \\
\hline
\end{tabular}%
}
\end{table}

We compare CPU, GPU, and FPGA implementations of the complete image-to-$F_z$ pipeline. The software baselines run on an AMD Ryzen 7 8845H CPU and an NVIDIA RTX 4060 Laptop GPU, while the hardware implementation targets a Xilinx Zynq UltraScale+ MPSoC (XCZU19EG-2FFVC1760I). As shown in Table~\ref{tab:edge_deployment_performance}, the FPGA completes the pipeline in 1.09~ms with an energy cost of 8.41~mJ/frame, compared with 6.82~ms and 238.05~mJ/frame on the CPU and 3.26~ms and 33.60~mJ/frame on the GPU, while maintaining comparable force accuracy.

Dynamic performance was evaluated using the vibration platform shown in Fig.~S1 of the supplementary material. As shown in Fig.~\ref{fig:vibration_frequency_response}, all platforms recover the 50-Hz excitation. At 70~Hz, the CPU exhibits aliasing, whereas the GPU and FPGA retain the correct response. At 100~Hz, only the FPGA preserves both the spectral peak and temporal periodicity. These results demonstrate the advantage of FPGA deployment for deterministic high-bandwidth tactile processing. Detailed per-stage benchmarks, energy measurements, output rates, bandwidth estimates, discrete Fourier transform (DFT) processing, and measured aliasing frequencies are provided in Section~III of the supplementary material. Additional demonstrations of 3D reconstruction, force feedback grasping, and vibration measurement are provided in Supplementary Video~S1.

\section{Conclusion}
FasTac is a compact curved tactile sensor integrating RGB-NIR imaging, boundary-prior 3D reconstruction, three-axis force estimation, and FPGA acceleration. It achieves a depth MAE of 0.0415~mm, while HyperForce obtains NMAE values of 2.74\% and 2.39\% for normal and shear forces, respectively. The FPGA image-to-$F_z$ pipeline runs in 1.09~ms per frame. Multi-object reconstruction, feedback grasping, and vibration measurements validate high-precision geometry perception, stable force feedback, and high-speed dynamic sensing.

\bibliographystyle{IEEEtran}
\bibliography{references}

\end{document}

\maketitle

\renewcommand{\thefigure}{S\arabic{figure}}
\renewcommand{\thetable}{S\arabic{table}}

\section{Data Collection and Training Protocol}

In the setup shown in Fig.~6 of the main paper, FasTac was fixed on a CNC platform, and a spherical indenter mounted on an ATI Nano17 force/torque sensor applied controlled contacts. Tactile images, CNC poses, and force readings were recorded synchronously. Four corner points defined the valid sensing region, within which calibration grids were generated by bilinear interpolation.

At each grid point, vertical contacts were recorded under normal loads of 0.5, 1.0, and 1.5~N. Under each normal preload, tangential loads of $F_h=0.3F_z$ and $F_h=0.6F_z$ were applied in four in-plane directions. Real-time $F_z$ feedback adjusted the CNC Z-axis position to maintain stable contact. Vertical samples were used to supervise surface-normal and depth estimation, whereas all valid normal and tangential samples were used for force and displacement estimation. Pixel-wise normal and depth ground truth was rendered from the CNC pose, indenter radius, and aligned sensor CAD model.

Spatially disjoint grids were used to prevent neighboring contact locations from appearing in different subsets. The training, validation, and test grids had sizes of $8\times8$, $5\times5$, and $8\times8$, yielding 192, 75, and 192 geometry samples and 1504, 569, and 1497 force samples, respectively.

The normal-estimation MLP was trained using Adam with an initial learning rate of $10^{-3}$ and a batch size of 32 for up to 300 epochs. HyperForce was trained using AdamW with an initial learning rate of $10^{-4}$, a weight decay of $10^{-3}$, and a batch size of 8 for up to 900 epochs.

\section{High-Frequency Vibration Measurement Platform}

\begin{figure}[t]
  \centering
  \includegraphics[width=\linewidth]{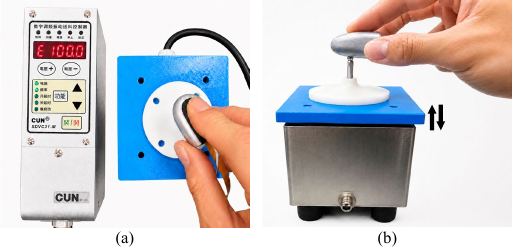}
  \caption{\textbf{High-frequency vibration measurement platform.} (a) Frequency controller and top view of the platform. (b) Side view of the platform.}
  \label{fig:supp_vibration_setup}
\end{figure}

As shown in Fig.~\ref{fig:supp_vibration_setup}, the frequency controller drives the platform in sinusoidal vertical motion, causing the probe to periodically indent the FasTac surface and produce a dynamic normal-force response.

\section{Edge Deployment and Dynamic Evaluation Details}

The CPU and GPU baselines were evaluated on an AMD Ryzen 7 8845H CPU and an NVIDIA RTX 4060 Laptop GPU, respectively, while the FPGA implementation targeted a Xilinx Zynq UltraScale+ MPSoC (XCZU19EG-2FFVC1760I). FasTac was connected through an FPC cable to an FMC-to-MIPI adapter mounted on the FPGA board.

\begin{table}[t]
\centering
\caption{Detailed Performance of CPU, GPU, and FPGA Implementations on $128\times128$ Inputs}
\label{tab:supp_edge_deployment_performance}
\scriptsize
\setlength{\tabcolsep}{1.5pt}
\renewcommand{\arraystretch}{1.15}
\resizebox{\columnwidth}{!}{%
\begin{tabular}{@{} c c c c c @{}}
\hline
\textbf{Scope} & \textbf{Platform} & \textbf{Latency (ms)} & \textbf{Accuracy (MAE)} & \textbf{Energy (mJ/frame)} \\
\hline
\multirow{3}{*}{\makecell{Surface normal\\estimation}}
& CPU & 2.40  & \textbf{0.0456} mm/pixel & 74.83 \\
& GPU & 0.30  & \textbf{0.0456} mm/pixel & 3.00 \\
& FPGA & \textbf{0.11} & 0.0508 mm/pixel & \textbf{0.29} \\
\hline
\multirow{3}{*}{\makecell{Depth\\reconstruction}}
& CPU & 1.17 & \textbf{0.0496} mm & 21.51 \\
& GPU & 1.34 & \textbf{0.0496} mm & 17.28 \\
& FPGA & \textbf{0.87} & 0.0741 mm & \textbf{8.23} \\
\hline
\multirow{3}{*}{\makecell{Normal-force\\estimation}}
& CPU & 2.30 & \textbf{0.0668} N & 74.15 \\
& GPU & 0.36 & \textbf{0.0668} N & 11.01 \\
& FPGA & \textbf{0.11} & \textbf{0.0668} N & \textbf{0.18} \\
\hline
\multirow{3}{*}{\makecell{Image-to-\\normal-force}}
& CPU & 6.82 & 0.0669 N & 238.05 \\
& GPU & 3.26 & \textbf{0.0667} N & 33.60 \\
& FPGA & \textbf{1.09} & 0.0679 N & \textbf{8.41} \\
\hline
\end{tabular}%
}
\end{table}

For edge deployment, the original $1920\times1080$ images used in the offline experiments of the main paper were cropped and downsampled to $128\times128$. This preprocessing accommodates the limited FPGA buffering and computation resources, while the smaller sensor readout window increases the camera frame rate. Therefore, the accuracy values reported here differ from the full-resolution offline results because of the different input resolution and preprocessing. Table~\ref{tab:supp_edge_deployment_performance} reports the latency, accuracy, and energy of each processing stage and of the complete image-to-$F_z$ pipeline.

The energy values are reported per frame. CPU and GPU energy was obtained from platform power counters and divided by the number of processed frames, whereas FPGA energy was measured during the full-stream board run.

For dynamic evaluation, FasTac captured images at $f_s=240~\mathrm{Hz}$, while the CPU, GPU, and FPGA pipelines produced $F_z(t)$ outputs at approximately 104, 156, and 240~Hz, respectively. The aliasing-free bandwidth was approximated as $f_{\max}\approx\min(f_s,f_{\mathrm{out}})/2$, giving limits of about 52, 78, and 120~Hz. The DC component of each $F_z(t)$ sequence was removed, and the dominant nonzero DFT peak was selected as the measured vibration frequency.

At 50~Hz, all platforms recovered the correct spectral peak and excitation periodicity. At 70~Hz, the CPU aliased to 31.58~Hz, whereas the GPU and FPGA measured 69.98 and 70.03~Hz, respectively. At 100~Hz, the CPU and GPU peaks shifted to 1.10 and 49.39~Hz and no longer followed the excitation cycles, whereas the FPGA measured 100.04~Hz and retained the correct temporal periodicity.